\documentclass[letterpaper, 10 pt, conference]{ieeeconf}
\IEEEoverridecommandlockouts 
\usepackage[letterpaper, left=54pt, right=54pt, bottom=54pt, top=54pt]{geometry}

\usepackage{graphicx}
\usepackage{amsmath} 
\usepackage{amssymb}  
\usepackage{mathtools}
\usepackage{bbm}
\usepackage{dsfont}
\usepackage{multirow}
\usepackage[dvipsnames]{xcolor}
\usepackage{subfig}
\usepackage[noadjust]{cite}
\usepackage{enumerate}
\usepackage{booktabs}
\usepackage{hyperref}

\usepackage{algpseudocode}
\usepackage{algorithm}
\usepackage{caption}
\usepackage[free-standing-units=true]{siunitx}
\usepackage{cleveref}

\title{\LARGE \bf
Seamless Interaction Design with Coexistence and Cooperation Modes for Robust Human-Robot Collaboration}

\author{Zhe Huang*, Ye-Ji Mun*, Xiang Li$\dagger$, Yiqing Xie$\dagger$, Ninghan Zhong$\dagger$, \\ Weihang Liang, Junyi Geng, Tan Chen, and Katherine Driggs-Campbell
\thanks{* denotes equal contribution as the first author. $\dagger$ denotes equal contribution as the second author.}
\thanks{Z. Huang, Y. Mun, X. Li, Y. Xie, W. Liang, and K. Driggs-Campbell are with the Department of  Electrical and Computer Engineering at the University of Illinois at Urbana-Champaign. emails: \{zheh4, yejimun2, xiangl5, yiqingx2, weihang2, krdc\}@illinois.edu}
\thanks{N. Zhong is with the Department of Computer Science at the University of Illinois at Urbana-Champaign. email: ninghan2@illinois.edu}
\thanks{J. Geng is with the Robotics Institute at Carnegie Mellon University. email: junyigen@andrew.cmu.edu}
\thanks{T. Chen is with the Coordinated Science Laboratory at the University of Illinois at Urbana-Champaign. email: tanchen@illinois.edu}
\thanks{We thank Foxconn Interconnect Technology for supporting the authors with their research. This work was supported in part by ZJU-UIUC Joint Research Center Project  No. DREMES 202003, funded by Zhejiang University.}
}

\begin{document}

\maketitle
\thispagestyle{empty}
\pagestyle{empty}

\begin{abstract}
A robot needs multiple interaction modes to robustly collaborate with a human in complicated industrial tasks. We develop a Coexistence-and-Cooperation (CoCo) human-robot collaboration system. Coexistence mode enables the robot to work with the human on different sub-tasks independently in a shared space. Cooperation mode enables the robot to follow human guidance and recover failures. A human intention tracking algorithm takes in both human and robot motion measurements as input and provides a switch on the interaction modes. We demonstrate the effectiveness of CoCo system in a use case analogous to a real world multi-step assembly task. \end{abstract}
\section{Introduction}\label{sec:Introduction}

Human-robot collaboration is one of representative ongoing revolutions in Industry 4.0~\cite{robla2017working}. With barriers between human and robot removed, automation can be significantly improved by integrating robust cognition and flexible manipulation from human workers, and thus higher productivity can be achieved~\cite{wang2017trends}. Collaborative robotics solutions are actively entering the industrial market~\cite{villani2018survey}. In a multi-step assembly task~\cite{kuka2018bmw}, a human worker moves parts to corresponding fixtures and puts rivets to predefined locations, and then a robot installs the rivets. The human and the robot work in parallel to reduce assembly cycle time. The human works on pick-and-place alignment which requires dexterity, and the robot works on non-ergonomic rivet installation. Besides labor division, the robot solution includes key features such as safety enforcement by contact detection, repetitive trials for installation on misaligned rivets, and mode switch between autonomous and compliant mode by pressing buttons and detecting interaction forces.



Though this tailored collaborative robot solution presents impressive performance gain, there is still huge potential of further improvement in more flexible manufacturing scenarios. Contact detection works after the contact accident between human and robot has already happened. The contact accident can be prevented if vision is applied to detect human hand position. The assumption of assembly failure recovery from several trials is the failure can be effectively detected and the misalignment is sufficiently mild to be addressed by repeated attempts. During the autonomous mode, the human needs a large amount of practice to coordinate with the pre-defined robot motion plan. A fixed trajectory plan does not fit a more free-form task, where the order of task steps is determined by the human herself. A more natural mode switch mechanism is desired to substitute additional operations such as pressing buttons. We introduce a Coexistence-and-Cooperation (CoCo) human-robot collaboration system to address these potential limitations, and demonstrate the effectiveness in a more flexible multi-step assembly task.  
\section{Method}\label{sec:Method}

\begin{figure}[t]
    \centering
    \includegraphics[width=\linewidth]{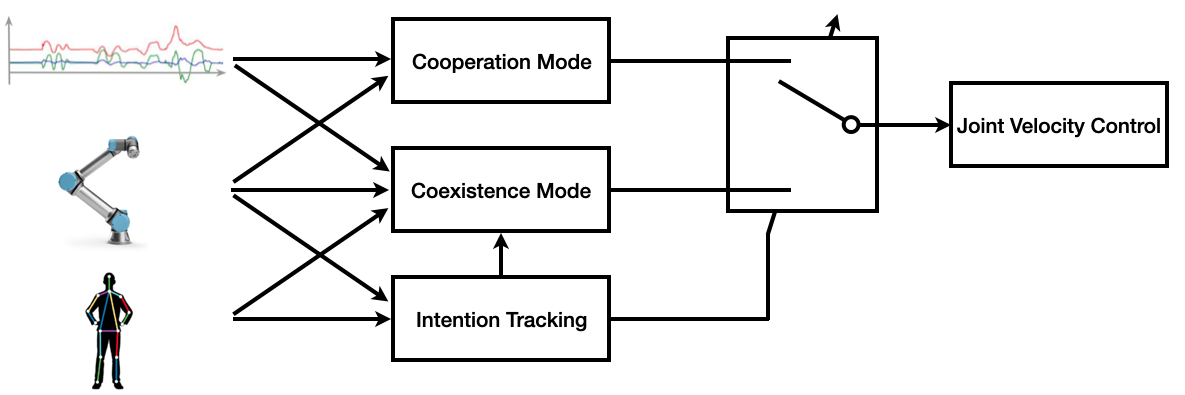}
    \caption{Overview of Coexistence-and-Collaboration (CoCo) human-robot collaboration system.}
    \label{fig:CoCo}
    \vspace{-10pt}
\end{figure}

The Coexistence-and-Collaboration (CoCo) human-robot collaboration system follows a labor division regime similar to~\cite{kuka2018bmw} where the human aligns parts and the robot performs assembly actions. The architecture of CoCo is illustrated in Figure~\ref{fig:CoCo}. The inputs to the system comprise force-torque measurements, robot proprioception and human body poses. The inputs in our case study specifically are force measurements, robot end-effector position, and human right hand position. We use a UR5e robot as the collaborative robot, which contains a force-torque sensor on the wrist. Human body pose estimation is performed using OpenPose~\cite{cao2017realtime} on Intel RealSense RGBD cameras. There are two separate control modules representing Coexistence Mode and Cooperation Mode. During the Coexistence Mode, the robot is performing assembly actions while not interrupting the human working on part alignment. The robot and the human coexist in the shared work space, but they work on different tasks and do not desire disturbance from each other. During the Cooperation Mode, the robot allows to be manually guided by the human to reach to a designated location and perform assembly action in order to recover an assembly failure. Both outputs from Coexistence Mode and Cooperation Mode are Cartesian velocity commands of the end-effector. A mode switch is controlled by a human intention tracking module to choose the control command, which will be fed into inverse kinematics and sent to joint velocity controller. The human intention tracking module is inspired by Mutable Intention Filter~\cite{huang2021long}. The human hand position is used as input to track the desired goal region where the human aligns the part. Both the end-effector position and the human hand position are used as input to track whether the human wants to work without interruption or manually guide the robot. 

The Coexistence Mode follows the paradigm of artificial potential field~\cite{khatib1986real}. Based on the maximum likelihood estimate of the desired goal region, waypoints are created for the robot to follow to perform the assembly action. The current waypoint serves as the subgoal, which generates a goal-directed end-effector velocity command added with a repulsive velocity command from the human hand position. The main component of the Cooperation Mode is admittance control which introduces manual guidance. When the robot is initially at the Coexistence Mode and the intention tracking module identifies human wants the robot to switch to the Cooperation Mode, the end-effector receives an attractive velocity command towards human hand. When contact is detected, the velocity command is set as zero and admittance control is activated. The end of manual guidance is detected by zero contact, which will switch the Cooperation Mode back to the Coexistence Mode. The immediate waypoint is set below the current robot end-effector position for the robot to perform assembly action. Assembly action during the Coexistence Mode is monitored by wrench thresholding to ensure the assembly action is finished. Velocity smoothing is applied to smoothly transition between the modes.
\section{Case Study}\label{sec:case-study}
\begin{figure}[t]
    \centering
    \includegraphics[width=\linewidth]{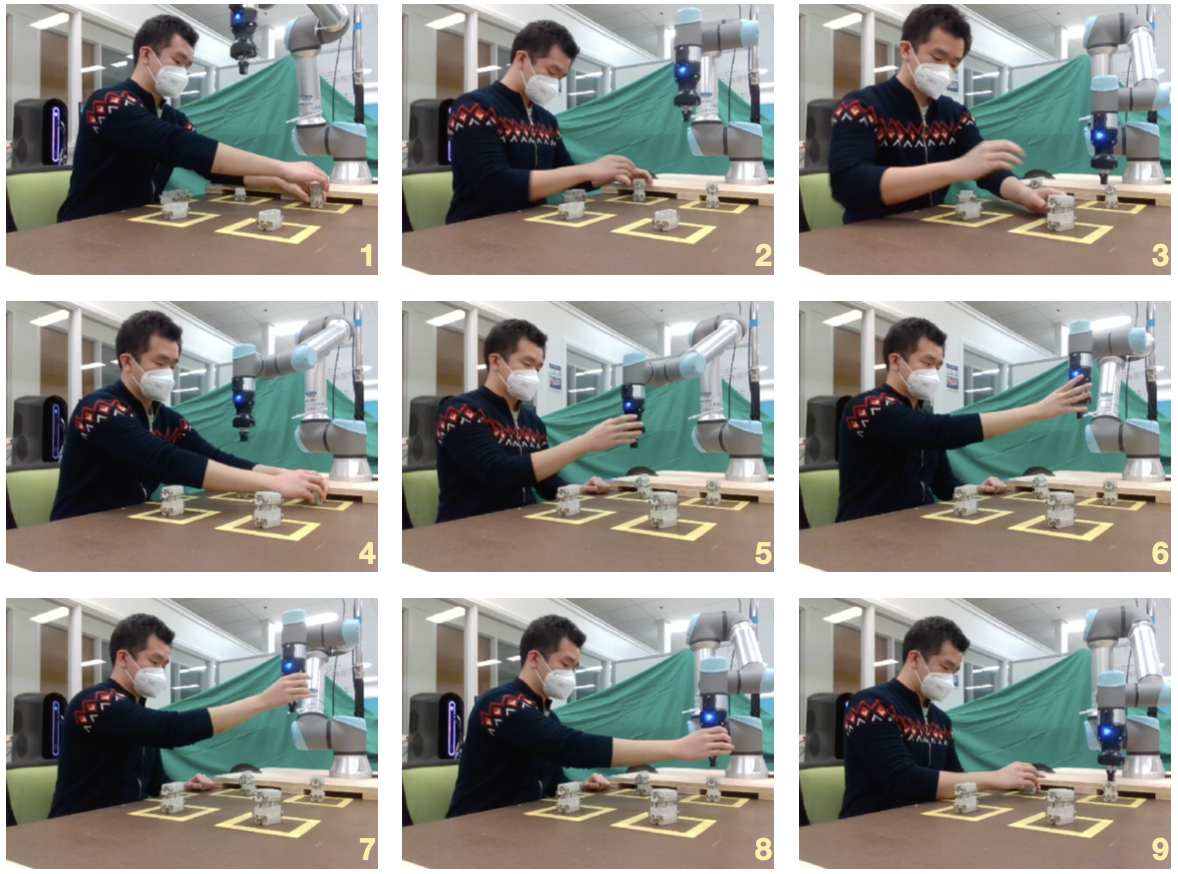}
    \caption{Demonstration on a free-form multi-step assembly use case.}
    \label{fig:use-case}
    \vspace{-10pt}
\end{figure}

We demonstrate CoCo system in a multi-step assembly task which is similar to the task in~\cite{kuka2018bmw} but designed in a more free-form way. The human-robot team needs to assemble four pairs of parts, where female parts are separately located at four regions, and male parts are in a preparation area which can be easily reached by the human. The human decides the order of part alignment which is unknown to the robot. The robot needs to perform pushing action on aligned part pairs to accomplish the assembly action. Figure \ref{fig:use-case} shows a trial of the assembly task. In this trial, the human followed the task execution sequence of top left $\rightarrow$ bottom left $\rightarrow$ top right $\rightarrow$ bottom right to align the parts. Initialized with the Coexistence Mode, the robot identified the first pair of aligned parts was at top left, and performed pushing action as presented in Figure \ref{fig:use-case}.3, but the assembly was failed due to an appropriate pushing point. The human noticed the assembly failure, and recovered the part alignment in Figure~\ref{fig:use-case}.4. The human then reached out to the end-effector and attempted to guide the robot for failure recovery. The robot was at the Coexistence Mode, so the end-effector moved away from the human hand as shown in Figure~\ref{fig:use-case}.5-6. After the cooperation intention of human was identified, the robot moved towards the human hand, and activated manual guidance after detecting intentional human robot contact as presented in Figure~\ref{fig:use-case}.7-8. After the manual guidance was finished, the robot performed assembly action at the location determined by the human, and recovered the assembly failure as presented in Figure~\ref{fig:use-case}.9.
\section{Conclusions and Future Work}\label{sec:conclusions}

We implement a Coexistence-and-Cooperation system for seamless and robust human-robot collaboration in free-form multi-step assembly tasks. We demonstrate that the system uses vision-based approaches to avoid contact accidents in Coexistence Mode. The robust recovery is realized by letting human take the lead in detecting and recovering assembly failure during Cooperation Mode. Intention tracking allows the robot to coordinate with human task execution and not vice versa. Natural mode switch is achieved without additional button pressing operations. Our future work includes conducting human subject experiments and quantifying the performance of the system. We will implement an ablation study to evaluate respective contribution of each component to the overall system.

\bibliographystyle{IEEEtran}
\bibliography{bib}

\begin{thebibliography}{1}
\providecommand{\url}[1]{#1}
\csname url@rmstyle\endcsname
\providecommand{\newblock}{\relax}
\providecommand{\bibinfo}[2]{#2}
\providecommand\BIBentrySTDinterwordspacing{\spaceskip=0pt\relax}
\providecommand\BIBentryALTinterwordstretchfactor{4}
\providecommand\BIBentryALTinterwordspacing{\spaceskip=\fontdimen2\font plus
\BIBentryALTinterwordstretchfactor\fontdimen3\font minus
  \fontdimen4\font\relax}
\providecommand\BIBforeignlanguage[2]{{%
\expandafter\ifx\csname l@#1\endcsname\relax
\typeout{** WARNING: IEEEtran.bst: No hyphenation pattern has been}%
\typeout{** loaded for the language `#1'. Using the pattern for}%
\typeout{** the default language instead.}%
\else
\language=\csname l@#1\endcsname
\fi
#2}}

\bibitem{robla2017working}
S.~Robla-G{\'o}mez, V.~M. Becerra, J.~R. Llata, E.~Gonzalez-Sarabia,
  C.~Torre-Ferrero, and J.~Perez-Oria, ``Working together: A review on safe
  human-robot collaboration in industrial environments,'' \emph{IEEE Access},
  vol.~5, pp. 26\,754--26\,773, 2017.

\bibitem{wang2017trends}
Y.~Wang and F.~Zhang, \emph{Trends in control and decision-making for
  human-robot collaboration systems}.\hskip 1em plus 0.5em minus 0.4em\relax
  Springer, 2017.

\bibitem{villani2018survey}
V.~Villani, F.~Pini, F.~Leali, and C.~Secchi, ``Survey on human--robot
  collaboration in industrial settings: Safety, intuitive interfaces and
  applications,'' \emph{Mechatronics}, vol.~55, pp. 248--266, 2018.

\bibitem{kuka2018bmw}
{KUKA - Robots \& Automation}, ``Innovative human-robot collaboration for
  bmw/mini crash can assembly,'' \url{https://youtu.be/keh99z1M5LI}, 2018.

\bibitem{cao2017realtime}
Z.~Cao, T.~Simon, S.-E. Wei, and Y.~Sheikh, ``Realtime multi-person 2d pose
  estimation using part affinity fields,'' in \emph{Proceedings of the IEEE
  conference on computer vision and pattern recognition}, 2017, pp. 7291--7299.

\bibitem{huang2021long}
Z.~{Huang}, A.~{Hasan}, K.~{Shin}, R.~{Li}, and K.~{Driggs-Campbell},
  ``Long-term pedestrian trajectory prediction using mutable intention filter
  and warp lstm,'' \emph{IEEE Robotics and Automation Letters}, vol.~6, no.~2,
  pp. 542--549, 2021.

\bibitem{khatib1986real}
O.~Khatib, ``Real-time obstacle avoidance for manipulators and mobile robots,''
  in \emph{Autonomous robot vehicles}.\hskip 1em plus 0.5em minus 0.4em\relax
  Springer, 1986, pp. 396--404.

\end{thebibliography}

\end{document}